\documentclass[sigconf]{acmart}

\setcopyright{none}                         % 不打印版权信息
\renewcommand\footnotetextcopyrightpermission[1]{} % 去掉第一页左下角那块 permission 文本

\AtBeginDocument{%
  }

\setcopyright{acmlicensed}
\copyrightyear{2018}
\acmYear{2018}
\acmDOI{XXXXXXX.XXXXXXX}
\acmConference[Conference acronym 'XX]{Make sure to enter the correct
  conference title from your rights confirmation email}{June 03--05,
  2018}{Woodstock, NY}
\acmISBN{978-1-4503-XXXX-X/2018/06}

\usepackage{enumitem}
\usepackage{multirow}
\usepackage{soul}

\setlist[itemize]{leftmargin=*}
\begin{document}

\fancyhead{}  % 再保险一次，清空页眉
\title{DenoiseSplat: Feed-Forward Gaussian Splatting for Noisy 3D Scene Reconstruction}

\author{Fuzhen Jiang}
\authornote{Corresponding author.}

\email{*23060111@hdu.edu.cn}

\affiliation{%
  \institution{Hangzhou Dianzi University}
  \city{Hangzhou}
  \state{Zhejiang}
  \country{China}
}

\author{Zhuoran Li}
\email{23062207@hdu.edu.cn}
\affiliation{%
  \institution{Hangzhou Dianzi University}
  \city{Hangzhou}
  \state{Zhejiang}
  \country{China}
}

\author{Yilin Zhang}
\email{13959886169@163.com}
\affiliation{%
  \institution{Hangzhou Dianzi University}
  \city{Hangzhou}
  \state{Zhejiang}
  \country{China}
}

\renewcommand{\shortauthors}{Trovato et al.}


\begin{abstract}
3D scene reconstruction and novel-view synthesis are fundamental for VR, robotics, and content creation.
However, most NeRF and 3D Gaussian Splatting pipelines assume clean inputs and degrade under real noise and artifacts.
We therefore propose \textbf{\emph{DenoiseSplat}}, a feed-forward 3D Gaussian splatting method for noisy multi-view images.
We build a large-scale, scene-consistent noisy--clean benchmark on RE10K by injecting Gaussian, Poisson, speckle, and salt-and-pepper noise with controlled intensities.
With a lightweight MVSplat-style feed-forward backbone, we train end-to-end using only clean 2D renderings as supervision and no 3D ground truth.
On noisy RE10K, DenoiseSplat outperforms vanilla MVSplat and a strong two-stage baseline (IDF + MVSplat) in PSNR/SSIM and LPIPS across noise types and levels.
\end{abstract}

\begin{CCSXML}
<ccs2012>
   <concept>
       <concept_id>10010147.10010178.10010224.10010226.10010239</concept_id>
       <concept_desc>Computing methodologies~3D imaging</concept_desc>
       <concept_significance>500</concept_significance>
       </concept>
 </ccs2012>
\end{CCSXML}

\ccsdesc[500]{Computing methodologies~3D imaging}

\keywords{3D scene reconstruction, 3D Gaussian splatting, noisy multi-view inputs, image denoising, neural rendering}



\maketitle

\begin{figure*}
    \centering
    \includegraphics[width=1\textwidth, trim=0 40 0 5, clip]{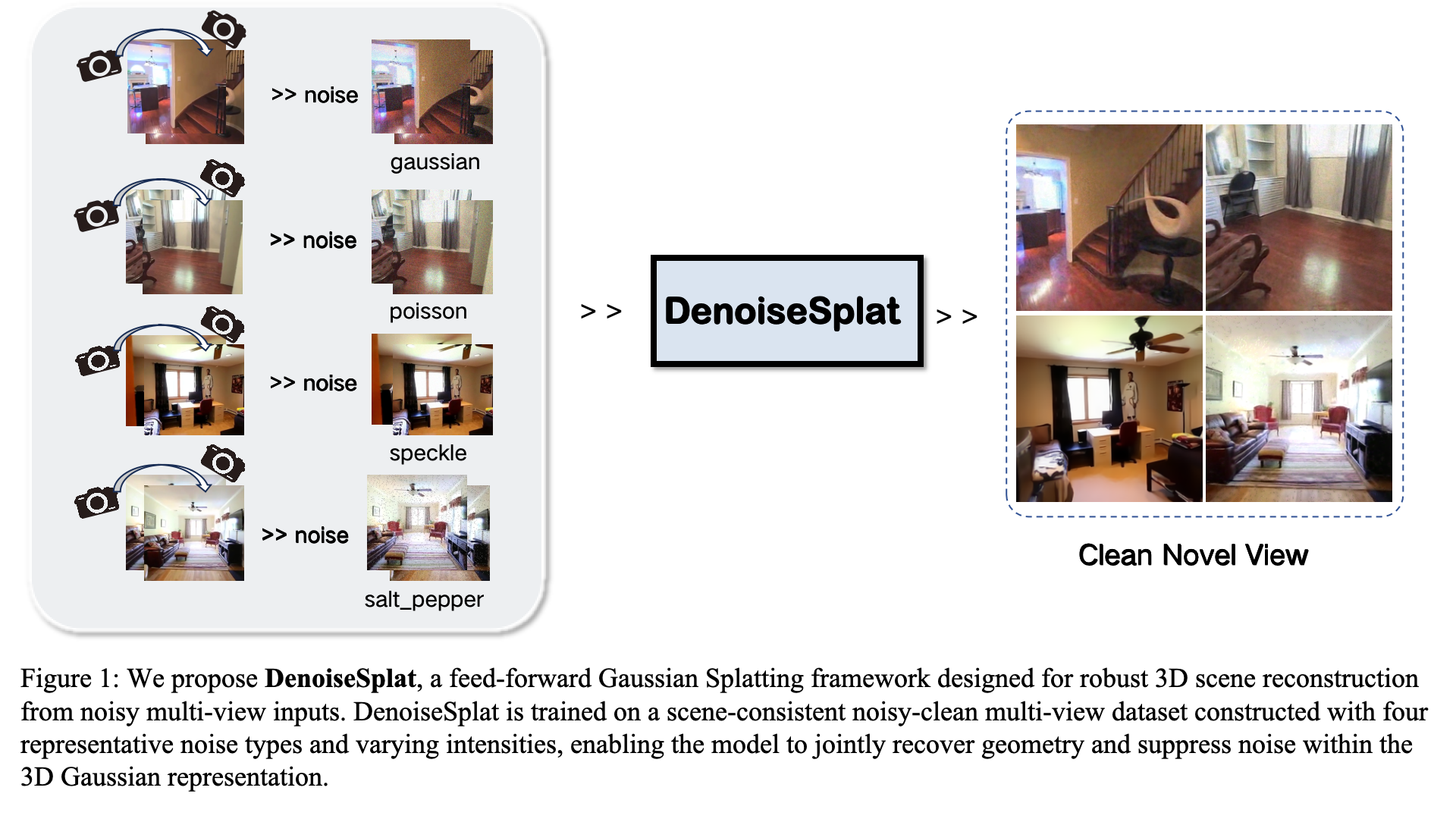}
    \label{fig:teaser}
\end{figure*}

\section{Introduction}
\label{sec:intro}

3D reconstruction and novel view synthesis are core capabilities for applications such as virtual reality, robotic navigation, and digital content creation. Neural Radiance Fields (NeRF)~\cite{mildenhall2020nerf} and its follow-ups have achieved remarkable progress in novel view synthesis and high-fidelity 3D scene reconstruction, albeit at the cost of expensive per-scene optimization. More recently, 3D Gaussian Splatting (3DGS)~\cite{kerbl3Dgaussians} represents scenes as a set of anisotropic 3D Gaussians and achieves real-time rendering with high visual quality. Building on this idea, feed-forward frameworks such as MVSplat~\cite{chen2024mvsplat} directly predict 3D Gaussian primitives from sparse multi-view images without per-scene optimization, enabling efficient reconstruction from casually captured videos.

However, most of these methods implicitly assume that the input images are \emph{clean}: captured under well-controlled conditions or carefully preprocessed, with negligible noise or degradation. In contrast, real-world multi-view data---for example, clips from web videos or mobile devices---often suffer from sensor noise, compression artifacts, and low-light degradation. Noise not only corrupts geometry estimation but also damages textures and fine details, especially when fusing information across sparse views. Under this realistic setting, 3D reconstruction from noisy multi-view inputs remains largely under-explored in the current NeRF/3DGS literature.

A straightforward engineering solution is a two-stage pipeline. One first denoises each frame in the 2D image domain using modern learning-based denoisers, such as the recent generalizable IDF network~\cite{kim2025idf} or classical CNN-based models like DnCNN~\cite{zhang2017dncnn} and FFDNet~\cite{zhang2018ffdnet}, and then feeds the denoised images into NeRF- or 3DGS-based pipelines for 3D reconstruction. While simple and modular, this design has notable drawbacks: (1) denoising can easily introduce over-smoothing and remove subtle details; (2) independently denoising each view weakens multi-view consistency, which is crucial for reliable 3D fusion; and (3) decomposing the pipeline into multiple modules increases inference latency and system complexity. At the other extreme, directly feeding noisy images into vanilla 3DGS or MVSplat leads to rapid degradation of reconstruction quality as noise intensity increases, as we will empirically demonstrate in Sec.~\ref{sec:experiments}.

In this work, we instead tackle noise \emph{inside the 3D representation}. Rather than treating denoising as an external pre-processing step, we aim to let a feed-forward 3D Gaussian splatting network itself learn to recover clean 3D scenes from noisy multi-view observations. Concretely, we build on the multi-view encoding and 3D decoding architecture of MVSplat and propose \emph{DenoiseSplat}, a feed-forward 3D Gaussian splatting method tailored to noisy inputs. Given noisy images and camera parameters as input and clean renderings as the only supervision, DenoiseSplat is trained end-to-end to predict a clean 3D Gaussian representation and its rendered views, without any access to ground-truth 3D supervision.

Fig.~1 offers a teaser overview of our works. To support this setting, we construct a multi-noise, scene-consistent noisy--clean dataset based on RealEstate10K (RE10K). For each scene, we inject one of four noise types---Gaussian, Poisson, speckle, or salt-and-pepper---into the 2D RGB images and keep the noise type and level consistent across all views within that scene, yielding paired noisy--clean multi-view samples. This simple yet controlled degradation pipeline mimics realistic acquisition conditions with scene-level noise characteristics and provides a reproducible benchmark for studying noisy 3D reconstruction.

To further improve robustness under heavy noise, we introduce a \emph{dual-branch Gaussian head} for geometry--appearance decoupling, as illustrated in Fig.~2. Instead of using a single shared head to regress all Gaussian parameters, we decouple geometry-related and appearance-related predictions into two lightweight branches: a geometric branch that outputs centers, rotations, scales, and opacities, and an appearance branch that predicts spherical-harmonics coefficients and colors. This design allows geometry to be estimated from more stable, noise-robust cues, while appearance is refined separately to absorb residual noise and color fluctuations. At test time, given only noisy multi-view inputs, a single forward pass through DenoiseSplat produces a denoised 3DGS scene and high-quality novel-view renderings, without any test-time optimization.

Our main contributions are summarized as follows:
\begin{itemize}[leftmargin=*,
                topsep=0pt,itemsep=0pt,parsep=0pt,partopsep=0pt]
    \item \textbf{Problem formulation and feed-forward framework for noisy multi-view reconstruction.}
    We systematically study 3D reconstruction from noisy multi-view inputs and, building on MVSplat, propose DenoiseSplat, a feed-forward 3D Gaussian splatting framework specifically designed for noise-corrupted images. Our method preserves the advantages of no per-scene optimization and efficient rendering, while explicitly enhancing robustness to noise through architectural and training choices.
    
    \item \textbf{Dual-branch Gaussian head for geometry--appearance decoupling.}
    We redesign the Gaussian head into a dual-branch structure, where a geometric branch predicts position-, rotation-, scale-, and opacity-related parameters and an appearance branch predicts spherical-harmonics coefficients and colors. This geometry--appearance decoupling mitigates the interference between noisy appearance and stable geometry, leading to more consistent 3D structure and sharper textures under strong noise.
    
    \item \textbf{Multi-noise, scene-consistent noisy--clean data construction.}
    We develop a simple yet reproducible noise degradation pipeline on RE10K, introducing four noise types (Gaussian, Poisson, speckle, and salt-and-pepper) and adopting a scene-level noise configuration, where all views of a scene share the same noise type and level. This setting better reflects realistic acquisition conditions and provides a useful benchmark for studying noisy 3D reconstruction.
    
    \item \textbf{Comprehensive experiments and ablations on noise levels and design choices.}
    On the constructed noisy RE10K benchmark, we compare DenoiseSplat against vanilla MVSplat and strong two-stage baselines that combine state-of-the-art 2D denoisers with 3D reconstruction under multiple noise types and intensities. We further conduct ablation studies on the standard deviation of Gaussian noise, the hyperparameters of other noise types, and the proposed dual-branch Gaussian head, together with qualitative analyses under different noise settings. Across a wide range of noise regimes, our method consistently outperforms baselines in PSNR, SSIM, and LPIPS, while preserving more texture details and geometric consistency.
\end{itemize}

\section{Related Works}
\label{sec:related}

\subsection{Multi-view 3D Reconstruction and Gaussian Splatting Representations}

Classical multi-view reconstruction methods rely on geometric techniques such as structure-from-motion (SfM), multi-view stereo (MVS), and bundle adjustment to recover scene geometry via feature matching and triangulation. While effective in many settings, their performance degrades in the presence of sparse textures, strong noise, or severe occlusions. NeRF~\cite{mildenhall2020nerf} represents scenes as implicit neural radiance fields and employs differentiable volumetric rendering to achieve high-quality novel view synthesis. Subsequent works have made substantial progress in accelerating training and inference and in improving rendering quality, yet the cost of volumetric rendering remains relatively high.

3D Gaussian Splatting (3DGS)~\cite{kerbl3Dgaussians} instead represents the scene explicitly as a set of Gaussian primitives, each parameterized by position, covariance, opacity, and color, and uses rasterization-based rendering for efficient image synthesis. This formulation strikes a favorable balance between reconstruction quality and computational efficiency. Methods built on 3DGS have been successfully applied to few-view reconstruction, super-resolution, language-driven editing, and 3D segmentation, among other tasks. For example, MMGS improves sparse-view synthesis through multi-model synergy, SRSplat studies feed-forward super-resolution from sparse low-resolution multi-view inputs, REALM enables open-world reasoning, segmentation, and editing directly on Gaussian scenes, and Sparse4DGS extends Gaussian splatting to sparse-frame dynamic scene reconstruction \cite{shi2025mmgs,hu2025srsplat,shi2025realm,shi2025sparse4dgs}. Feed-forward frameworks that directly predict Gaussian primitives from multi-view images via volumetric feature aggregation further eliminate per-scene optimization, enabling a single forward pass mapping from multi-view images to a 3DGS representation.

However, this entire line of work is largely designed and evaluated under clean-input assumptions, where image quality is high and noise is negligible. There is still a lack of systematic analysis and targeted modeling for realistic acquisition conditions, in which noisy measurements, diverse degradations, and cross-view consistency issues are ubiquitous.

\subsection{Image Denoising and Noise-Robust Vision Models}

Image denoising is a long-standing problem in low-level vision. Classical methods such as non-local means and BM3D~\cite{dabov2007bm3d} explicitly exploit image statistics and non-local self-similarity priors, and remain competitive for simple additive Gaussian noise. With the rise of deep learning, supervised denoising networks such as DnCNN~\cite{zhang2017dncnn}, FFDNet~\cite{zhang2018ffdnet}, and RIDNet~\cite{anwar2019ridnet} trained on synthetic noisy--clean pairs have demonstrated superior performance on standard benchmarks.

However, many CNN-based denoisers are trained for a fixed noise type or a narrow range of noise levels, and often struggle to generalize to unseen degradations or mixed noise. 
Recently, IDF (Iterative Dynamic Filtering)~\cite{kim2025idf} proposes a compact yet powerful denoising network that generates pixel-wise dynamic kernels and applies them iteratively, achieving strong generalization to diverse noise types and levels despite being trained only on single-level Gaussian noise. 
In this work, we adopt IDF as a strong 2D denoising expert in our Denoise-Then-MVSplat baseline due to its efficiency and generalization ability, and compare it against our proposed 3D noise-aware reconstruction pipeline.

\subsection{Robust 3D Reconstruction from Degraded Inputs}

For other forms of degradation such as blur, compression, or weather effects, several works have explored explicitly modeling input degradation in 3D reconstruction and novel view synthesis. Examples include jointly learning deblurring, deblocking, or deraining modules within NeRF-like frameworks, as well as introducing uncertainty modeling and robust estimation for dynamic or challenging scenes. These studies collectively indicate that tightly coupling degradation modeling with 3D representations can improve reconstruction quality and robustness.

Nevertheless, \emph{dedicated studies on 3D reconstruction from noisy multi-view inputs remain limited}. On the one hand, most existing methods either assume preprocessed inputs or do not systematically account for different noise types and levels. On the other hand, some robust approaches rely on per-scene optimization or complex inference procedures, which limits their scalability and applicability to real-time or large-scale scenarios.

Several recent works aim to make NeRF- or 3DGS-based methods more robust to degraded inputs. 
Deblur-NeRF~\cite{ma2022deblurnerf} explicitly models spatially-varying blur kernels to recover sharp radiance fields from defocus or motion-blurred images. 
More recently, RobustGS~\cite{wu2025robustgs} introduces a degradation-aware feature enhancement module that can be plugged into feed-forward 3DGS pipelines to better handle low-quality inputs such as noise, low light, or rain. 
These methods demonstrate that explicitly modeling degradations in the reconstruction pipeline can substantially improve robustness. 
However, they typically focus on specific degradations (e.g., blur) or treat ``low-quality'' conditions in a more generic way, 
and few works systematically study multi-type, multi-level additive noise on large-scale multi-view datasets. 
In contrast, our work targets noisy multi-view reconstruction with diverse noise types and intensities, 
and provides a dedicated noisy--clean benchmark together with a feed-forward 3D Gaussian Splatting network trained end-to-end on noisy inputs.

\section{Methodology}
\label{sec:method}

\begin{figure*}[t]
  \centering
  \includegraphics[width=1\linewidth, trim=0 170 0 55, clip]{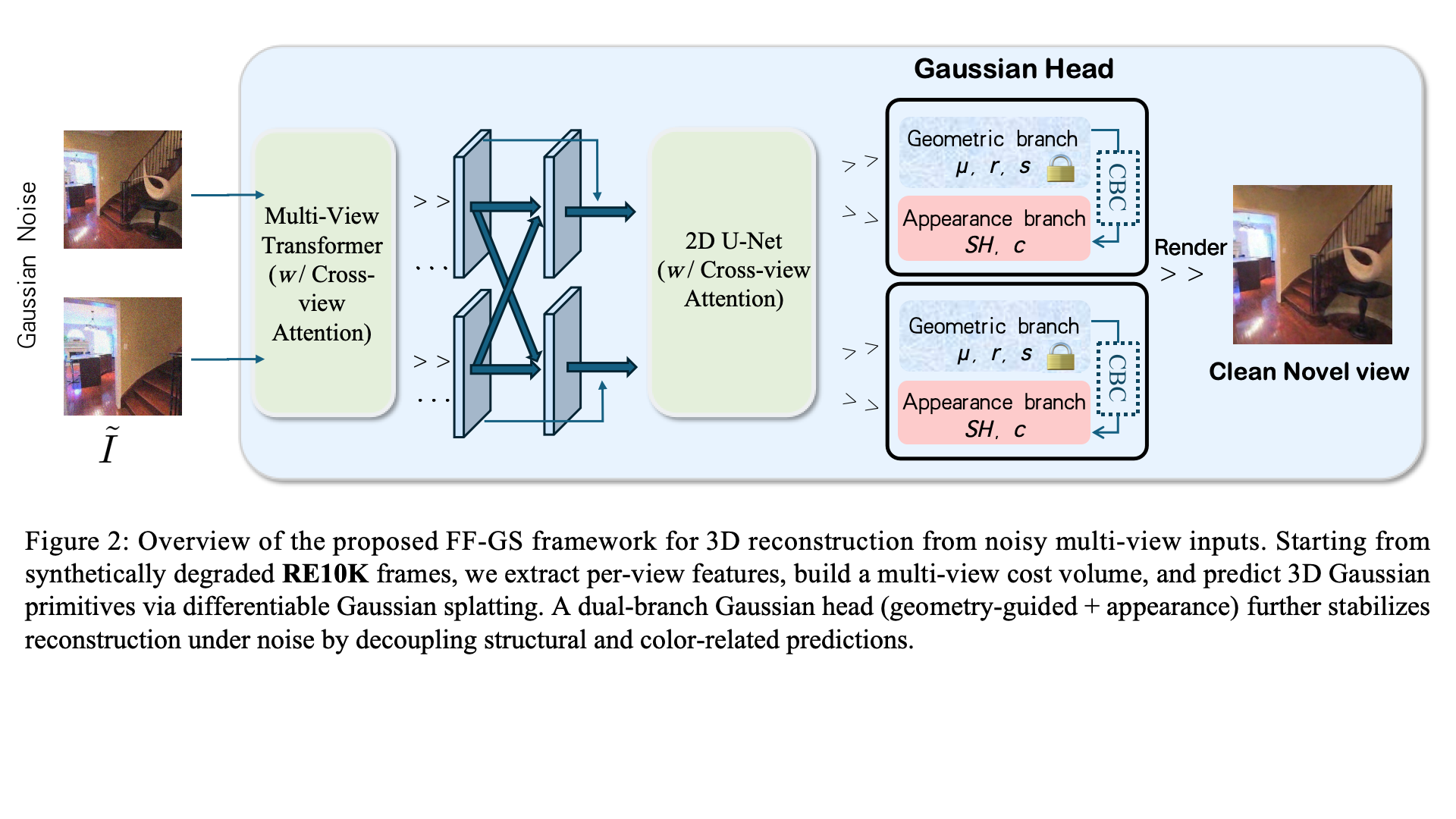}

  \label{fig:pipeline}
\end{figure*}

Our goal is to directly predict a clean 3D Gaussian scene representation and its renderings from noisy multi-view inputs. The overall pipeline is illustrated in Fig.~2. We first construct multi-noise-type noisy--clean paired multi-view data based on RE10K. Then, we design a feed-forward Gaussian splatting network on top of the MVSplat framework, which takes noisy images as input and is trained end-to-end with clean renderings as supervision. At test time, a single forward pass suffices to obtain a clean 3DGS scene and high-quality novel views.

\subsection{Problem Definition and Overall Pipeline}
\label{subsec:problem_overview}

Given a collection of 3D scenes
\(
\mathcal{S} = \{s\}
\),
each scene \(s\) provides \(V\) clean images
\(
\{I^{\text{GT}}_{s,v}\}_{v=1}^{V}
\)
with corresponding camera parameters
\(
\{\mathcal{C}_{s,v}\}_{v=1}^{V}
\).
Following the noise degradation process in Sec.~\ref{subsec:noise_degradation_dataset}, we synthetically generate noisy observations by applying a 2D RGB-domain corruption operator \(\mathcal{D}(\cdot)\) to each clean image while keeping the viewpoints unchanged:
\begin{equation}
\bigl\{\tilde{I}^{\text{noise}}_{s,v},\, I^{\text{GT}}_{s,v},\, \mathcal{C}_{s,v}\bigr\}_{v=1}^{V},
\qquad
\tilde{I}^{\text{noise}}_{s,v}=\mathcal{D}\!\left(I^{\text{GT}}_{s,v}\right).
\end{equation}
Here \(\tilde{I}^{\text{noise}}_{s,v}\) denotes the noisy image obtained in the 2D RGB domain; the camera parameters and underlying scene geometry are shared with the corresponding clean view by construction.

Our goal is to learn parameters \(\theta\) of a feed-forward predictor that maps multi-view noisy observations (and cameras) to a 3D Gaussian scene representation and its renderings:
\begin{equation}
f_{\theta}:\ 
\bigl(\{\tilde{I}^{\text{noise}}_{s,v}\}_{v=1}^{V}, \{\mathcal{C}_{s,v}\}_{v=1}^{V}\bigr)
\;\longrightarrow\;
\bigl(\mathcal{G}_{s}, \{\hat{I}_{s,v}\}_{v=1}^{V}\bigr),
\end{equation}
where \(\mathcal{G}_{s}\) is the reconstructed 3D scene represented by Gaussian primitives, and \(\hat{I}_{s,v}\) is rendered from \(\mathcal{G}_{s}\) under \(\mathcal{C}_{s,v}\).
During training, we use only 2D image-domain supervision by comparing \(\hat{I}_{s,v}\) with \(I^{\text{GT}}_{s,v}\).
At test time, the clean images are not available; the model predicts \(\mathcal{G}_{s}\) and \(\hat{I}_{s,v}\) \emph{conditioned on} noisy multi-view inputs and camera parameters in a single forward pass.

\subsection{Noise Degradation and Dataset Construction}
\label{subsec:noise_degradation_dataset}

\textbf{Modeling multiple noise types.}
We build our noisy multi-view benchmark on top of the RealEstate10K (RE10K) dataset~\cite{realestate10k}, which contains camera trajectories for approximately 80{,}000 video clips collected from around 10{,}000 YouTube real-estate videos, 
totalling about 10 million frames. For each clip, the dataset provides a sequence of calibrated camera poses along a viewing trajectory, making it a standard testbed for novel view synthesis and 3D scene reconstruction. Starting from the original clean multi-view RE10K data, we synthesize four common noise types in the 2D RGB image domain:
Gaussian noise, Poisson noise, speckle noise, and salt-and-pepper noise. Let
\(
I \in [0,1]^{H\times W \times 3}
\)
denote an input normalized image and \(\tilde{I}\) the degraded one. The four noise types can be summarized as:
\begin{itemize}
    \item Gaussian noise: \(\tilde{I} = \mathrm{clip}(I + \mathcal{N}(0,\sigma^{2}))\);
    \item Poisson noise: \(\tilde{I} = \mathrm{clip}\bigl(s \cdot \mathrm{Poisson}(I/s)\bigr)\);
    \item Speckle noise: \(\tilde{I} = \mathrm{clip}\bigl(I + I \odot \mathcal{N}(0,\sigma^{2})\bigr)\);
    \item Salt-and-pepper noise: each pixel is set to 0 or 1 with probability \(\alpha\).
\end{itemize}
Here \(\sigma\) or \(\alpha\) is uniformly sampled within a predefined range, e.g., the standard deviation of Gaussian noise is sampled from \([0.08, 0.12]\), and the total ratio of salt-and-pepper noise is sampled from \([0.015, 0.03]\). All specific values are managed through a unified configuration file in our implementation.

\noindent \textbf{Scene-level noise configuration.}
In practice, images from the same scene are typically captured with the same device and exposure settings, so their noise types and strengths tend to be consistent across views. To reflect this, we adopt a scene-level noise sampling strategy. For each scene \(s\):
\begin{enumerate}
    \item uniformly sample a noise type \(t_{s}\) from the four candidates;
    \item sample a noise parameter \(\phi_{s}\) within the intensity range corresponding to type \(t_{s}\);
    \item apply the same pair \((t_{s}, \phi_{s})\) to all views \(\{v\}\) of that scene.
\end{enumerate}
The resulting noisy--clean multi-view data thus maintains noise consistency at the scene level while exhibiting diversity across scenes. All noise configurations (types and strengths) are recorded in a separate metadata file to facilitate reproducibility and ablation studies.

\subsection{Feed-forward Gaussian Splatting Network}
\label{subsec:ff_gs_network}

\noindent \textbf{Gaussian scene representation and rendering.}
We follow 3DGS~\cite{kerbl3Dgaussians} and represent a scene \(s\) as a set of Gaussian primitives
\begin{equation}
\mathcal{G}_{s} = \{g_{i}\}_{i=1}^{N},\quad
g_{i} = (\mathbf{x}_{i}, \Sigma_{i}, \alpha_{i}, \mathbf{c}_{i}),
\end{equation}
where \(\mathbf{x}_{i} \in \mathbb{R}^{3}\) is the center position, \(\Sigma_{i} \in \mathbb{R}^{3\times 3}\) is the covariance matrix, \(\alpha_{i} \in [0,1]\) is the opacity, and \(\mathbf{c}_{i}\) denotes the appearance parameters (spherical-harmonics coefficients plus a base color).
Given camera parameters \(\mathcal{C}_{s,v}\) for view \(v\), a differentiable Gaussian splatting renderer \(\mathcal{R}\) produces
\begin{equation}
\hat{I}_{s,v} = \mathcal{R}(\mathcal{G}_{s}, \mathcal{C}_{s,v}) \in \mathbb{R}^{H\times W\times 3}.
\end{equation}

\noindent \textbf{MVSplat-based feed-forward architecture.}
Our network is built on the feed-forward design of MVSplat~\cite{chen2024mvsplat}: a shared multi-view 2D encoder extracts per-view features, which are then lifted and aggregated in a geometry-aligned manner to form a unified 3D representation (Fig.~2).
Concretely, the pipeline consists of three stages:

\begin{itemize}
    \item \textbf{Multi-view feature encoding.}
    For each noisy view \(\tilde{I}^{\text{noise}}_{s,v}\), a shared-weight 2D convolutional encoder produces multi-scale feature maps \(\{\mathbf{F}^{(l)}_{s,v}\}_{l=1}^{L}\).
    Rather than explicitly outputting a denoised image, the encoder learns feature representations that retain task-relevant structures while being tolerant to noise, under the downstream reconstruction objective.

    \item \textbf{Geometry-aligned 3D aggregation.}
    Conditioned on camera parameters \(\mathcal{C}_{s,v}\), we follow MVSplat and warp multi-view features into a plane-sweep volume via differentiable homography projection over discrete depth planes.
    The resulting 3D feature volume \(\mathbf{F}^{3\mathrm{D}}_{s}\) encodes multi-view consistency signals and serves as the input to the Gaussian prediction head.

    \item \textbf{Dual-branch Gaussian parameter prediction.}
    Different from the single shared head in MVSplat, we adopt a \emph{dual-branch Gaussian head} tailored for noisy inputs.
    The aggregated features \(\mathbf{F}^{3\mathrm{D}}_{s}\) are fed into two lightweight decoupled branches:
    (1) a \textbf{geometry branch} that predicts centers \(\mathbf{x}_{i}\), rotations, scales, and opacities \(\alpha_{i}\), and
    (2) an \textbf{appearance branch} that predicts spherical-harmonics coefficients and colors \(\mathbf{c}_{i}\).
    This decoupling is a modeling choice that allows the geometry branch to prioritize relatively stable structural cues, while the appearance branch can account for residual noise and color fluctuations.
    Empirically, this design improves structural consistency and texture sharpness across varying noise types and intensities.
\end{itemize}

\noindent \textbf{2D rendering supervision and loss functions.}
During training, we supervise reconstruction only in the 2D image domain by comparing \(\hat{I}_{s,v}\) with \(I^{\text{GT}}_{s,v}\).
We use a combination of pixel-wise \(\ell_{1}\) loss and structural similarity (SSIM):
\begin{equation}
\mathcal{L}
= \sum_{s}\sum_{v}
\bigl(
\lambda_{1}\,\lVert \hat{I}_{s,v} - I^{\text{GT}}_{s,v} \rVert_{1}
+ \lambda_{2}\,\bigl(1 - \mathrm{SSIM}(\hat{I}_{s,v}, I^{\text{GT}}_{s,v})\bigr)
\bigr),
\label{eq:reconstruction_loss}
\end{equation}
where \(\lambda_{1}\) and \(\lambda_{2}\) are weighting coefficients.
The loss is averaged over pixels within each view and summed over training scenes and views.

After end-to-end training on the noisy--clean paired data (Sec.~\ref{subsec:noise_degradation_dataset}), the model learns to predict a Gaussian scene representation whose renderings match the clean targets under the above objective.
At test time, given only noisy multi-view inputs, a single forward pass through DenoiseSplat outputs \(\mathcal{G}_{s}\) and corresponding renderings, without test-time optimization or an external 2D denoiser.

\subsection{Cross-Branch Boundary-Guided Appearance Correction (CBC)}
\label{subsec:cbc}

Our dual-branch design decouples geometry and appearance to improve robustness under noisy inputs, yet appearance estimation may still degrade when the geometry branch is uncertain, especially near geometric boundaries (e.g., depth/disparity discontinuities and occlusion edges).
In such regions, even small geometric inaccuracies can coincide with perceptually salient artifacts after rendering.
To mitigate this \emph{cross-branch error propagation} in a controlled manner, we introduce a lightweight \emph{cross-branch correction mechanism}, termed \textbf{CBC}, which leverages geometry-derived boundary strength and confidence as \emph{conditional signals} to gate a residual refinement in the appearance branch.

\noindent\textbf{Geometry-conditioned boundary gating.}
We construct a boundary-guided gating map from the full-resolution disparity (or depth) predicted by the geometry branch, together with a confidence proxy derived from the candidate distribution over disparity/depth hypotheses (e.g., obtained by applying \(\mathrm{softmax}\) to the cost-volume logits along the plane dimension).
Using disparity $\delta$ for clarity, we compute
\begin{equation}
E = \left\lVert \nabla \delta \right\rVert,\quad
C = \max(\mathrm{pdf}),\quad
B = \operatorname{norm}(E)\cdot(1-C),
\label{eq:cbc_gate}
\end{equation}
where $E$ denotes boundary strength (the gradient magnitude of $\delta$), and $C$ is the maximum probability of the disparity/depth candidate distribution $\mathrm{pdf}$ as a confidence proxy.
The resulting $B\in[0,1]$ emphasizes regions that are simultaneously boundary-like and low-confidence: $B$ increases when $E$ is strong and $C$ is small.

\noindent\textbf{Boundary-guided residual correction in appearance.}
Let $A$ denote the output of the appearance branch (e.g., SH/color-related parameters or logits).
CBC predicts a residual correction $\Delta A$ via a lightweight CNN $g(\cdot)$ conditioned on $(A,B,C)$:
\begin{equation}
\Delta A = g([A, B, C]),\quad
A' = A + B \odot \Delta A,
\label{eq:cbc_residual}
\end{equation}
where $[\cdot]$ denotes channel-wise concatenation and $\odot$ is the element-wise product.
This design applies refinement primarily where $B$ is large, thereby suppressing boundary artifacts while avoiding unnecessary modification in non-boundary or high-confidence regions.

\noindent\textbf{Implementation detail (gradient isolation).}
During training, we apply stop-gradient to the geometry-derived signals by detaching $B$ and $C$ (i.e., \texttt{detach}).
As a result, CBC updates only the appearance-branch parameters and the CBC module parameters, preventing appearance gradients from back-propagating into the geometry branch.
This realizes an explicit cross-branch interaction mechanism: geometry provides conditional boundary/confidence cues, and appearance performs a gated residual refinement accordingly.
We provide qualitative evidence of CBC in Fig.~3.

\begin{figure}[ht]
    \centering
    \label{cbc_zoom}
    \includegraphics[width=1\linewidth, trim=0 100 0 0, clip]{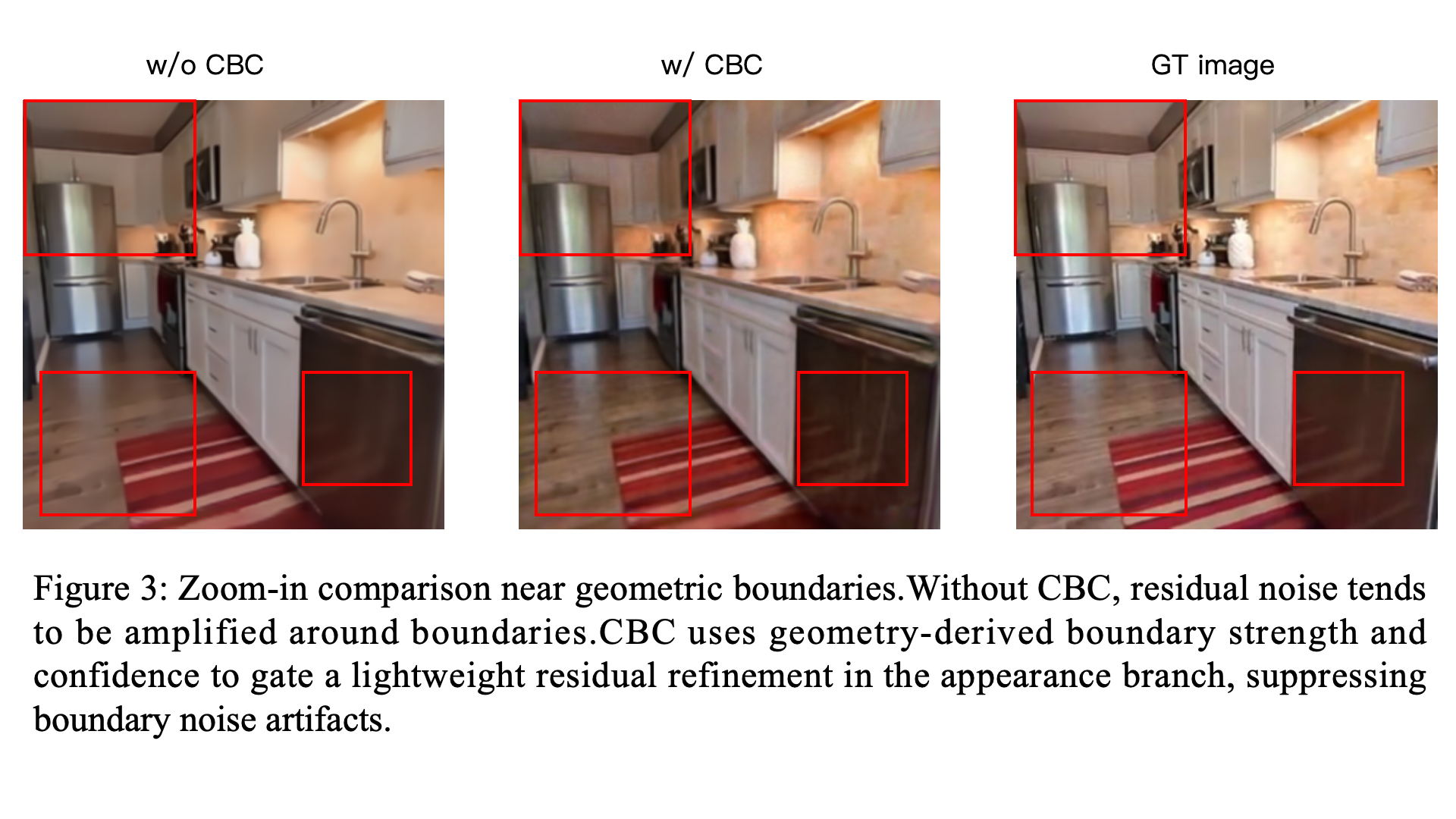}
\end{figure}

\section{Experiments}
\label{sec:experiments}

\begin{figure*}[t]
  \centering  \includegraphics[width=1\linewidth, trim=0 50 0 20, clip]{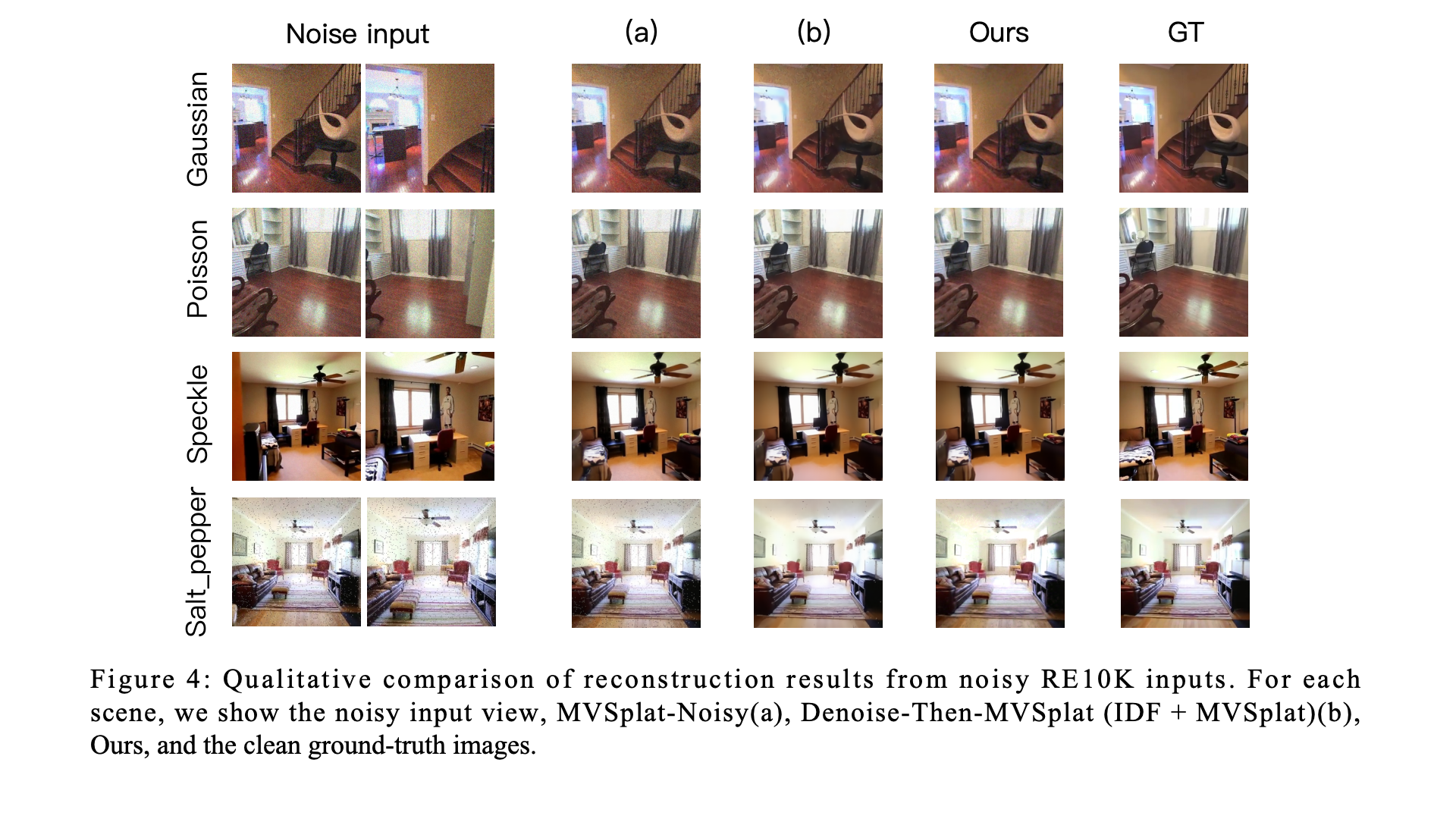}
  
  \label{fig:major_com_vis}
\end{figure*}

In this section, we evaluate the proposed method on the noisy multi-view dataset constructed from RE10K. We first describe the experimental setup and baselines, then present overall results and qualitative analyses, and finally conduct ablation studies to examine the effect of key design choices.

\subsection{Experimental Settings}
\label{subsec:exp_settings}

\noindent \textbf{Datasets and protocol.}
We use RealEstate10K (RE10K). For each scene, we uniformly sample a fixed number of frames as multi-view inputs and synthesize noisy--clean pairs as in Sec.~\ref{subsec:noise_degradation_dataset}.
Specifically, we apply a single randomly sampled noise type (Gaussian, Poisson, speckle, or salt-and-pepper) with a randomly sampled intensity (within the specified ranges) to \emph{all} views of the scene in the 2D RGB domain, producing \(\tilde{I}^{\text{noise}}_{s,v}\) with clean targets \(I^{\text{GT}}_{s,v}\).
Splits are scene-level with no overlap across train/val/test.
Unless stated otherwise, we report results on both \emph{seen views} (training views) and \emph{novel views} extrapolated along the camera trajectory, to assess reconstruction and generalization jointly.

\noindent \textbf{Metrics.}
We report PSNR (\(\uparrow\)), SSIM (\(\uparrow\)), and LPIPS (\(\downarrow\)), averaged over the test set; we additionally break down performance on seen vs.\ novel views.

\noindent \textbf{Baselines.}
We compare: (i) \textbf{MVSplat-GT} (upper bound), trained and evaluated on clean RE10K; 
(ii) \textbf{MVSplat-Noisy}, the clean-trained MVSplat evaluated directly on noisy inputs without fine-tuning;
(iii) \textbf{Denoise-Then-MVSplat}, a two-stage pipeline applying IDF~\cite{kim2025idf} (official pretrained weights, default inference) per view and then feeding the restored images into clean-trained MVSplat~\cite{chen2024mvsplat}, without fine-tuning either module; and
(iv) \textbf{Ours (DenoiseSplat)}, an MVSplat-style feed-forward 3DGS model retrained on our noisy--clean benchmark with noisy inputs and clean 2D supervision, without test-time optimization.

\noindent \textbf{Efficiency.}
To assess potential overhead, we report runtime/memory under a unified setting (single GPU, 256$\times$256, 2 context views).
We discard the first 5 batches for warm-up and report the mean runtime; results are summarized in Tab.~\ref{tab:efficiency}.

\begin{table}[ht]
\centering
    \small
    \caption{Inference efficiency comparison under the RE10K evaluation setting (256$\times$256, 2 context views). Encoder time is measured per scene, while decoder time is measured per rendered target view.}
    \begin{tabular}{lccccc}
    \toprule
    Method & Enc. (ms) & Dec. (ms) & Peak Mem. (GiB) \\
    \midrule
    MVSplat-Noisy & 58.12 & 1.92 & 1.03 \\
    Denoise-Then-MVSplat  & 86.90 & 1.92 & 1.28 \\
    \textbf{Ours} & 60.74 & 1.92 & 1.07 \\
    \bottomrule
    \end{tabular}
    \label{tab:efficiency}
\end{table}

\begin{table}
  \centering
  \caption{%
    Overall reconstruction performance on the noisy RE10K test set. 
    We report average PSNR$\uparrow$, SSIM$\uparrow$, and LPIPS$\downarrow$ over all test views.
  }
  \label{tab:overall_results}
  \begin{tabular}{lccc}
    \toprule
    Method & PSNR $\uparrow$ & SSIM $\uparrow$ & LPIPS $\downarrow$ \\
    \midrule
    MVSplat-GT(upper bound)    & 26.38 & 0.869 & 0.128 \\
    MVSplat-Noisy              & 24.46 & 0.702 & 0.349 \\
    Denoise-Then-MVSplat (IDF) & 24.77 & 0.788 & 0.272 \\
    \textbf{Ours}              & \textbf{25.05} & \textbf{0.814} & \textbf{0.260} \\
    \bottomrule
  \end{tabular}
\end{table}

\begin{table*}[t]
  \centering
  \caption{%
    Ablation on the standard deviation $\sigma$ of Gaussian noise for our DenoiseSplat. 
    We report average performance on the noisy RE10K test set.
  }
  \label{tab:gaussian_ablation}
\begin{tabular}{l|ccc|ccc|ccc}
    \multirow{2}{*}{Methods} & \multicolumn{3}{|c|}{MVSplat-Noisy} & \multicolumn{3}{c|}{Denoise-Then-MVSplat} & \multicolumn{3}{c}{\textbf{DenoiseSplat(Ours)}}\\
    \cline{2-10}
     & PSNR & SSIM & LPIPS  &
       PSNR & SSIM & LPIPS  &
       PSNR $\uparrow$ & SSIM $\uparrow$ & LPIPS $\downarrow$ \\
    \noalign{\hrule height 1.0pt}
    $\sigma$ = 0.05 & 25.19 & 0.777 & 0.298 & 25.28 & 0.801 & 0.256 & 25.44 & 0.814 & 0.243\\
    $\sigma$ = 0.08 & 24.62 & 0.700 & 0.365 & 24.82 & 0.732 & 0.300 & 24.86 & 0.784 & 0.282\\
    $\sigma$ = 0.12 & 23.54 & 0.612 & 0.432 & 24.26 & 0.665 & 0.396 & 24.33 & 0.742 & 0.328\\
    $\sigma$ = 0.15 & 22.79 & 0.557 & 0.472 & 23.14 & 0.603 & 0.436 & 23.87 & 0.712 & 0.360\\
\end{tabular}

\end{table*}

\subsection{Main Results}
\label{subsec:main_results}

\noindent \textbf{Overall performance.}
Tab.~\ref{tab:overall_results} summarizes the reconstruction accuracy and perceptual quality on the noisy RE10K test set.
Among methods that operate on noisy inputs, \textbf{DenoiseSplat} achieves the strongest overall trade-off across PSNR, SSIM, and LPIPS, and reduces the gap to the clean-input upper bound \textbf{MVSplat-GT}.
In contrast, directly feeding noisy images to \textbf{MVSplat-Noisy} leads to a pronounced drop, most notably in SSIM and LPIPS, which is consistent with texture blurring and local structural inconsistencies under noise.
Introducing a strong 2D denoiser in \textbf{Denoise-Then-MVSplat} improves all three metrics compared to \textbf{MVSplat-Noisy}.
Nevertheless, a noticeable margin to \textbf{MVSplat-GT} remains, especially in perceptual quality, suggesting that view-independent 2D restoration does not fully preserve the cues needed for multi-view reconstruction under challenging noise.

\noindent \textbf{Seen vs. novel views.}
Novel-view synthesis is generally more sensitive to geometric uncertainty than seen-view reconstruction.
Consistent with this, \textbf{MVSplat-Noisy} exhibits a larger performance drop on novel views, indicating difficulty in maintaining reliable multi-view consistency when the inputs are corrupted.
\textbf{Denoise-Then-MVSplat} narrows this gap to some extent, but its novel-view quality can still be limited when the 2D denoiser introduces view-dependent artifacts or removes fine details that are important for matching.
By comparison, \textbf{DenoiseSplat} maintains more stable performance across seen and novel views, with a smaller gap between them, which is indicative of improved cross-view coherence when noise is handled within the 3D reconstruction pipeline.

\noindent \textbf{Qualitative comparisons.}
Fig.~4 visualizes representative examples from the noisy RE10K test set.
On scenes with rich textures and complex boundaries, \textbf{MVSplat-Noisy} often leaves visible residual noise and exhibits color shifts or localized distortions.
\textbf{Denoise-Then-MVSplat} typically yields cleaner appearances, but frequently softens edges and suppresses high-frequency details, especially at higher noise levels.
We further highlight recurring \emph{failure patterns} of this two-stage pipeline in Fig.~5, including over-smoothing of fine textures, halo/ringing near strong boundaries, and erosion of thin structures (e.g., wires or railings).
These effects are consistent with view-independent 2D restoration, which may remove or alter high-frequency cues in a manner that is not fully consistent across views, thereby weakening multi-view correspondence and boundary fidelity.
In comparison, \textbf{DenoiseSplat} removes most visible noise while better preserving boundary sharpness and texture details, producing more coherent renderings with fewer artifacts in novel view synthesis.

\begin{figure}[ht]
  \centering
  \includegraphics[width=\linewidth, trim=0 200 0 140, clip]{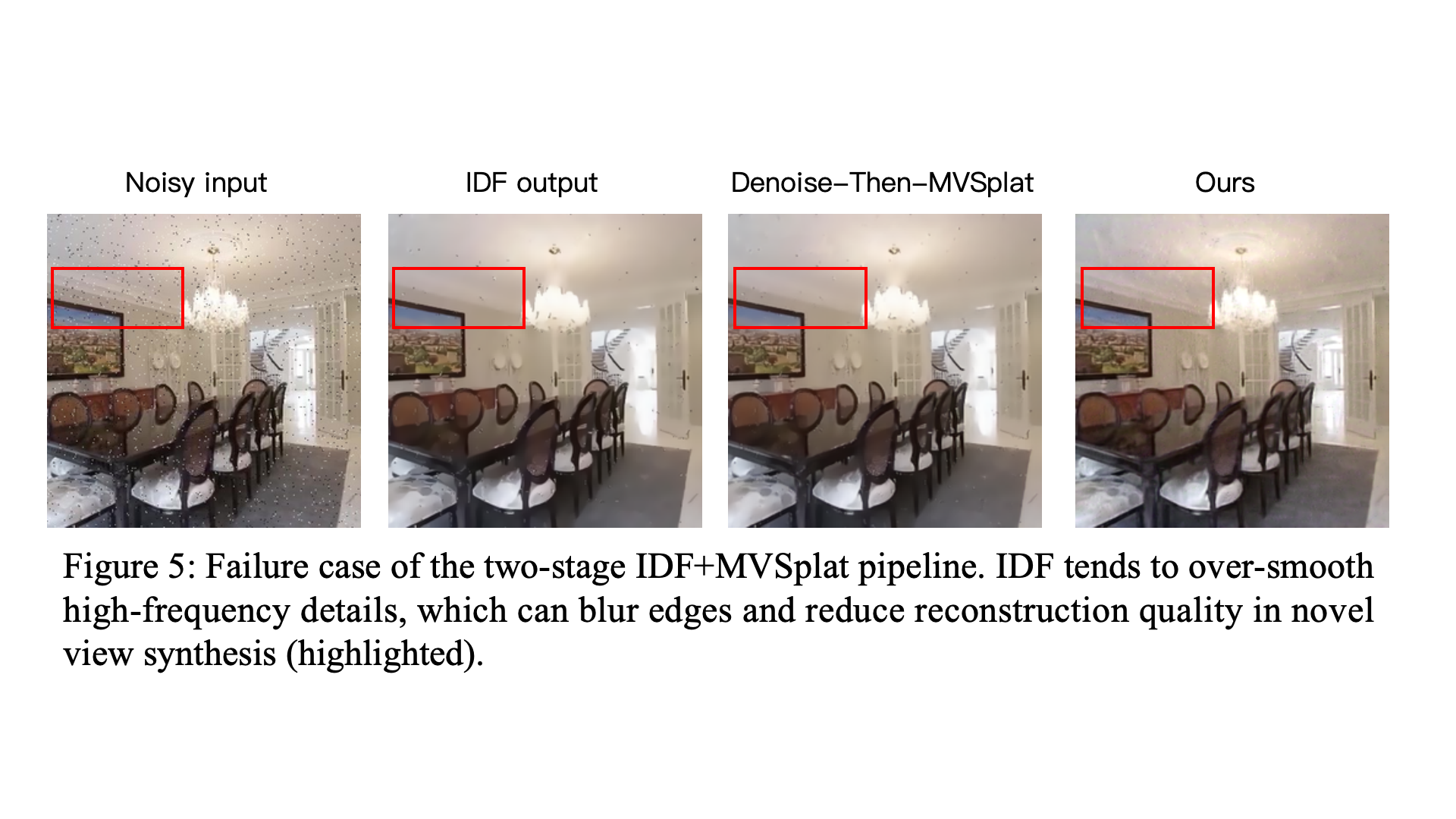}
  \label{fig:idf_failure}
\end{figure}

\subsection{Ablation Study}
\label{subsec:ablation}

We ablate robustness along the \emph{noise dimension}, varying noise \emph{type} and \emph{intensity} to examine trends in both metrics and visual artifacts.

\noindent \textbf{Effect of Gaussian noise level}
\label{subsubsec:ablation_gaussian_level}

We vary the Gaussian noise standard deviation
\(\sigma \in \{0.05,\ 0.08,\\\ 0.12,\ 0.15\}\)
and report results in Tab.~\ref{tab:gaussian_ablation}.
As \(\sigma\) increases, PSNR/SSIM decrease and LPIPS increases for all methods.
However, degradation rates differ: \textbf{MVSplat-Noisy} drops sharply for \(\sigma \ge 0.08\), while \textbf{Denoise-Then-MVSplat} remains competitive at low-to-medium noise but degrades faster at higher noise, consistent with front-end 2D denoising losing fine details required by downstream multi-view reconstruction.
In contrast, \textbf{DenoiseSplat} shows a smoother decline and maintains favorable LPIPS even under strong noise, indicating better generalization across noise intensities.

\noindent \textbf{Effect of noise level for other noise types}
\label{subsubsec:ablation_other_level}

We also vary noise hyperparameters for other types: Poisson ($scale$), speckle (multiplicative variance), and salt-and-pepper (\(\alpha \in \{0.01, 0.03\}\)); see Tab.~\ref{tab:other_noise_ablation}.

\begin{table}
  \centering
  \caption{%
    Ablation on the hyper-parameters of other noise types for our DenoiseSplat. 
    Each entry reports performances on the noisy RE10K test set.
  }
  \label{tab:other_noise_ablation}
  \begin{tabular}{lcccc}
    \toprule
    Noise type & Parameter setting & PSNR & SSIM & LPIPS \\
    \midrule
    \multirow{2}{*}{Poisson} & scale $ = 0.03$ & 24.10 & 0.740 & 0.331\\
                             & scale $ = 0.04$ & 23.74 & 0.723 & 0.350\\
    \midrule
    \multirow{2}{*}{Speckle} & $\sigma = 0.02$ & 25.43 & 0.846 & 0.177\\
                             & $\sigma = 0.05$ & 25.36 & 0.832 & 0.211\\
    \midrule
    \multirow{2}{*}{Salt--Pepper} & $\alpha = 0.01$ & 24.98 & 0.816 & 0.241\\
                                  & $\alpha = 0.03$ & 24.53 & 0.776 & 0.298\\
    \bottomrule
  \end{tabular}
\end{table}

\begin{figure*}[t]
  \centering
  \includegraphics[width=1\linewidth, trim=0 50 0 50, clip]{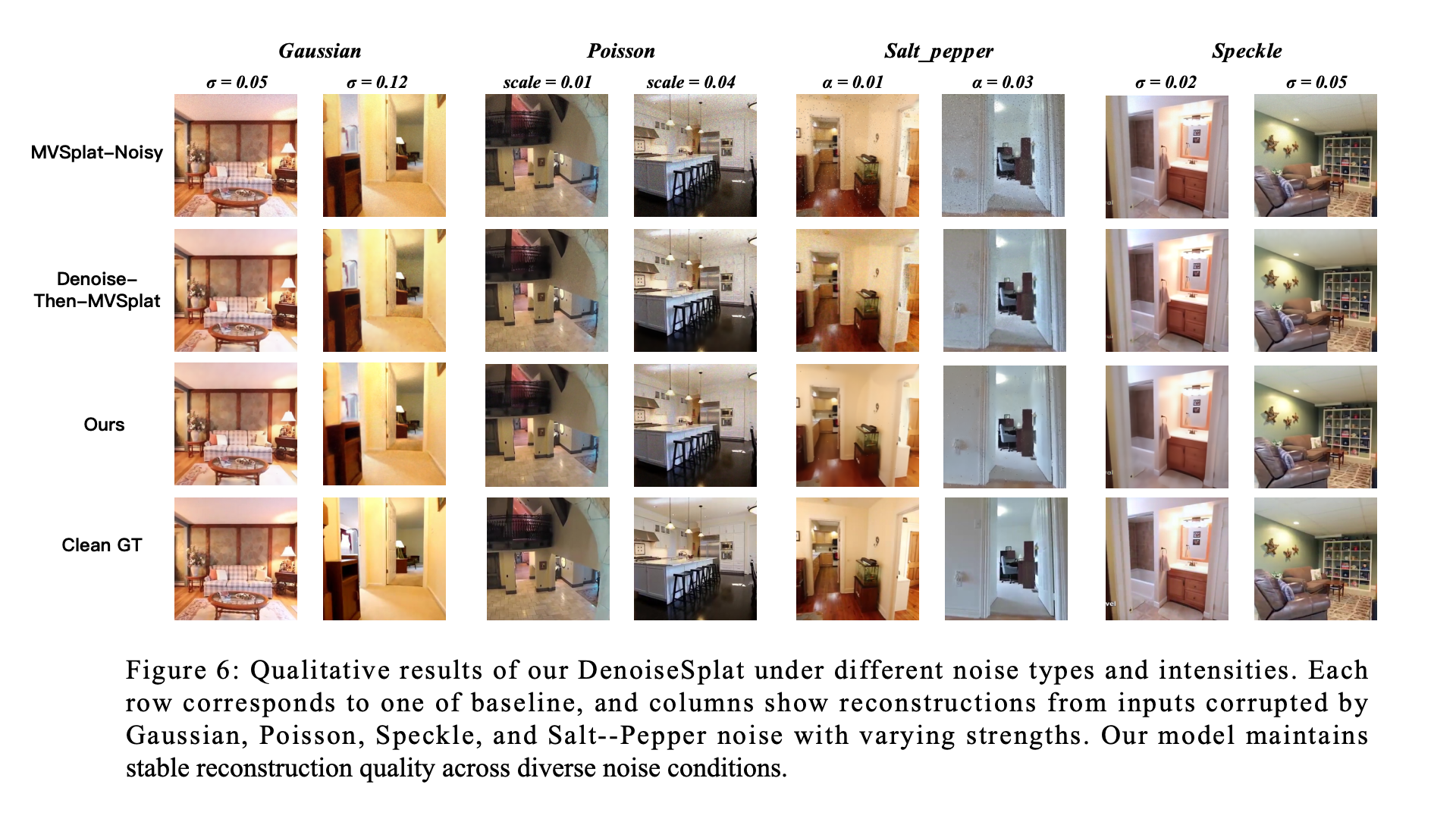}
  \label{fig:qualitative_noise}
\end{figure*}

We further include qualitative comparisons in Fig.~6 under representative settings (e.g., Gaussian \(\sigma=0.05\) vs.\ 0.12; Poisson $scale=0.03$ vs.\ 0.04; salt-and-pepper \(\alpha=0.01\) vs.\ 0.03; and speckle low vs.\ high variance).
At low noise, all methods recover main structures but \textbf{MVSplat-Noisy} often leaves residual noise; as noise increases, \textbf{MVSplat-Noisy} exhibits stronger artifacts and instability, \textbf{Denoise-Then-MVSplat} becomes smoother with weakened textures/boundaries, and \textbf{DenoiseSplat} better preserves contours and texture cues with more stable novel-view renderings.

\section{Conclusion}
\label{sec:conclusion}

In this paper, we studied 3D scene reconstruction from noisy multi-view inputs. Unlike classical NeRF and 3D Gaussian Splatting-based methods that assume clean input images, we focused on a setting that more closely reflects real-world acquisition conditions, where multiple noise types, varying noise intensities, and cross-view consistency all play important roles.

To address this problem, we proposed \emph{DenoiseSplat}, a feed-forward 3D Gaussian splatting network tailored to noisy inputs and built upon the MVSplat framework. We constructed a multi-noise, scene-consistent noisy--clean paired multi-view dataset on RealEstate10K by injecting Gaussian, Poisson, speckle, and salt-and-pepper noise into the 2D RGB domain, and trained DenoiseSplat end-to-end to map noisy multi-view images to clean renderings without accessing any ground-truth 3D supervision. Within this framework, a dual-branch Gaussian head decouples geometry- and appearance-related parameters, enabling the network to perform geometry reconstruction and noise suppression \emph{inside} the 3D representation. At test time, a single forward pass from noisy multi-view inputs yields a clean 3DGS scene and high-quality reconstructed views, without any test-time optimization.

On the constructed noisy RE10K benchmark, we compared DenoiseSplat against the original MVSplat and a strong two-stage baseline that combines a state-of-the-art 2D denoiser with MVSplat. Experimental results show that DenoiseSplat achieves consistently better or competitive PSNR, SSIM, and LPIPS across diverse noise types and intensities, while offering better texture preservation and geometric consistency in novel view synthesis. Ablation studies further confirm that jointly training under multiple noise types and varying noise intensities, together with the proposed dual-branch Gaussian head, is crucial for obtaining robust 3DGS representations under noisy inputs.

Despite these promising results, our work still has several limitations that merit further exploration. First, our current noise modeling is based mainly on synthetic noise and does not yet capture more complex degradations such as real camera noise, motion blur, and compression artifacts. Second, our experiments are primarily conducted on RE10K, and the cross-dataset and real-world generalization of the model remains to be systematically evaluated. Future work could incorporate real-noise captures or more accurate camera noise models, extend the framework to broader degradation types and dynamic scenes, and integrate our approach with higher-level semantic cues, aiming for feed-forward 3D representations that are not only noise-robust but also better aligned with downstream 3D understanding tasks.

\bibliographystyle{ACM-Reference-Format}
\bibliography{bib}

\end{document}